%% file: main.tex
\documentclass[mlhc,table]{jmlr}

\usepackage{longtable}

\usepackage{booktabs}
\usepackage{multirow} 
\usepackage{makecell} 
\usepackage[load-configurations=version-1]{siunitx} 
\usepackage{amsmath}
\usepackage{algorithm}
\usepackage{tikz}
\usepackage{wrapfig}
\usepackage[noend]{algpseudocode}
\usepackage{siunitx}

\theorembodyfont{\upshape}
\theoremheaderfont{\scshape}
\theorempostheader{:}
\theoremsep{\newline}

\jmlryear{2019}
\jmlrworkshop{Machine Learning for Healthcare, 2019} 

\def \Real{{\mathbb R}} 
\newcommand{\eValue}[1]{\mathbb{E}\left\{ #1 \right\}}

\newcommand{\ie}{i.e., }

\newcommand{\eg}{e.g. }

\newcommand{\iid}{i.i.d. }
\newcommand{\argmax}{\mathop{\mathrm{argmax}}}
\newcommand{\D}{\mathcal{D}}
\newcommand{\N}{\mathcal{N}}

\title[MGP and DCNN for Reconstruction of Reproductive Hormonal Dynamics]
{Multi-Task Gaussian Processes and Dilated Convolutional Networks for Reconstruction of Reproductive Hormonal Dynamics}

\author{\Name{I\~{n}igo Urteaga}\thanks{Contributed equally} \Email{inigo.urteaga@columbia.edu} \\
\addr{Department of Applied Physics and Applied Mathematics \\ Columbia University, New York City, NY, USA}\\
\Name{Tristan Bertin}\footnotemark[1]  \Email{tmb2183@columbia.edu} \\
\addr{Department of Biomedical Informatics\\ Columbia University, New York City, NY, USA} \\
\Name{Theresa M. Hardy} \Email{th89@nyu.edu} \\
\addr{Rory Meyers College of Nursing\\ New York University, New York City, NY, USA} \\
\Name{David J. Albers} \Email{david.albers@ucdenver.edu} \\
\addr{Division of Informatics, Department of Pediatrics \\ University of Colorado, Anschutz Medical Center, Denver, CO, USA}\\
\Name{No\'{e}mie Elhadad} \Email{noemie.elhadad@columbia.edu} \\
\addr{Department of Biomedical Informatics \\ Columbia University, New York City, NY, USA}
}

\begin{document}

\maketitle

\begin{abstract}
We present an end-to-end statistical framework for personalized, accurate, and minimally invasive modeling of female reproductive hormonal patterns. Reconstructing and forecasting the evolution of hormonal dynamics is a challenging task, but a critical one to improve general understanding of the menstrual cycle and personalized detection of potential health issues. Our goal is to infer and forecast individual hormone daily levels over time, while accommodating pragmatic and minimally invasive measurement settings. To that end, our approach combines the power of probabilistic generative models (i.e., multi-task Gaussian processes) with the flexibility of neural networks (i.e., a dilated convolutional architecture) to learn complex temporal mappings. To attain accurate hormone level reconstruction with as little data as possible, we propose a sampling mechanism for optimal reconstruction accuracy with limited sampling budget. Our results show the validity of our proposed hormonal dynamic modeling framework, as it provides accurate predictive performance across different realistic sampling budgets and outperforms baselines methods.
\end{abstract}
\vspace{0.5cm}
\section{Introduction}
\label{sec:intro}
\input{intro_significance}

\section{Dataset}
\label{sec:dataset}
\input{dataset}

\section{Methods}
\label{sec:methods}
\input{methods}

\section{Evaluation}
\label{sec:evaluation}
\input{evaluation} 

\section{Discussion and Related Work} 
\input{discussion}

\acks{We thank the anonymous reviewers for their feedback and comments, as well as ~\citet{clue} by BioWink GmbH for the information on cycle length and ovulation day. This work is supported in part by awards from the National Science Foundation (\#1344668) and the National Library of Medicine (R01 LM 912734).}

\clearpage
\input{main.bbl} 

\clearpage
\appendix
%
\section*{Evaluation}
\label{sec:evaluation_appendix}
\input{evaluation_appendix}

\end{document}

%% file: intro_significance.tex
We propose and validate a statistical framework for personalized, accurate, and minimally invasive modeling of female reproductive hormonal patterns. Starting from sparse hormone measurements at few and specific days, we aim to infer and predict daily hormone measurements over time. 
This is a challenging task, because there is much variation from one individual to another, both in the timing of the different phases of the menstrual cycle, and of the daily hormone levels during each phase~\citep{treloar1967variation,j-Alliende2002}.
As such, models that can adjust to individuals' patterns are needed; at the same time, population-level dynamics can also provide additional guidance towards our goal. Previous work was limited to modeling reproductive hormone levels of a given individual through independent Gaussian processes, and explored the impact of hormone measurement sampling frequency and sampling times on the ability to detect menstrual phases for that individual~\citep{j-Urteaga2017}.  
Here, we present a novel framework which departs from previous work in several ways, provides further understanding of the dynamics of the menstrual cycle, and produces generalizable insights for reconstruction and prediction of continuous temporal signals.

\subsection{Clinical Relevance}
The menstrual cycle reflects underlying female reproductive hormonal function. It can be described according to its length: \ie number of days from the first day of bleeding until the day before the next bleeding period~\citep{j-Hornsby2007,j-Small2007}, its phases (\eg follicular, luteal), and its daily reproductive hormonal dynamics, \eg estrogen ($E$), progesterone ($P$), inhibin (Ih), follicle stimulating hormone (FSH), and luteinizing hormone (LH). The menstrual cycle is a powerful indicator of overall health in women \citep{j-Small2007, j-Vassena2014}. Beyond its relevance to reproductive health and fertility~\citep{j-Filiberto2012,j-Vassena2014}, cycle characteristics can inform risk for chronic diseases such as cardiovascular disease, osteoporosis and cancer~\citep{bedford_prospective_2010,zittermann_physiologic_2000, solomon_menstrual_2002,shuster_premature_2010,mahoney_shift_2010}. Further, knowledge of an individual's full hormone dynamics can pinpoint hormonal irregularities and indicate potential diagnoses~\citep{j-Carmina1999,j-Giudice2010,tworoger201320}. In fact, the medical field has suggested menstruation should be considered ``the fifth vital sign''~\citep{acog2015vital,bobel_beyond_2019}.

Despite this realization, reproductive female physiology and the exact relationship between hormone levels and hormonal dynamics are still enigmatic in many ways. Due to challenges associated with timed hormone sample collection and the cost of repeated sampling~\citep{j-Filiberto2012}, continuous hormone measurement across the menstrual cycle has not been conducted on a large scale.

Currently, real-world population-level data is mostly available for cycle length as recorded through menstrual trackers. For a small subset of these populations, some phase information is also available through reports of ovulation test results. While these self-reported data have been validated~\citep{j-Hornsby2007,j-Small2007}, there are no real-world, large-scale datasets of hormonal dynamics throughout the menstrual cycle. Rather, our current understanding of continuous reproductive hormone levels is based on small-scale studies, \eg 30 women in~\citet{j-McLachlan1990}, and mechanistic models validated against these small datasets~\citep{j-Selgrade1999}. How to infer the reproductive hormonal dynamics of a given individual in a minimally-invasive, low-cost fashion is an open research question.

In this work, we contribute to \textbf{knowledge of the menstrual cycle} through real-world, self-tracked data obtained through a popular menstrual tracking app for menstrual cycle length and ovulation. This mitigates the potential limitations of mechanistic models which have been validated in small-scale studies only. Further, we provide and validate a method to \textbf{reconstruct and predict an individual's menstrual cycle} (\ie daily hormonal measurements, and thus phases and length) based on a few hormone measurements only. This data-driven approach to reconstructing an individual's cycle patterns can help with minimally-invasive, low-cost data collection at a massive scale.

\subsection{Technical Significance}
Mechanistic models of hormonal levels have been proposed in the literature~\citep{j-Selgrade1999,j-Clark2003,j-Selgrade2009}. These constitute a system of non-linear differential equations that describe the dynamics of the cycle, their physiology, and the inter-relations of the different reproductive hormones, and can be tuned through a set of parameters to generate different types of hormonal cycles. However, there are a number of limitations with these models. While they can simulate realistic cycles, additional machinery is needed for the inverse problem, namely inferring the full hormonal curve from few measurements.
Further, joint estimation of model states and parameters in a way that synchronizes the model to an individual, via techniques such as data assimilation~\citep{j-albers2018} is challenging. Accurate inference and avoidance of identifiability problems require measuring all the hormones at multiple time points; prerequisites currently not possible on a large scale~\citep{j-albers2018}. Previous hormone reconstruction work, which opted for a physiology agnostic approach to inferring hormonal cycles, experimented with individual Gaussian processes per-hormone for a single individual's data, where each hormone's dynamics were independently predicted~\citep{j-Urteaga2017}. 

The technical significance of this work is two-fold, as we contribute both from a data and a methods perspective. From a data standpoint, \textbf{(1)} we propose a method to generate synthetic, yet realistic datasets of full female hormonal cycles. The characteristics of these cycles are diverse, but grounded on real-world evidence: we anchor the high level cycle characteristics (namely cycle length and ovulation day) to those of a large, real-world dataset of self-reports.
From a modeling standpoint, our technical contributions are as follows. Since the overall aim is to devise models which necessitate as little data as possible, both in the number of individuals needed and the number of their hormonal measurements, we devise an end-to-end statistical framework for personalized modeling of female reproductive hormonal patterns under realistic sampling budgets: \textbf{(2)} We combine the power of generative processes with the ability of neural networks to learn complex temporal mappings. To this end, we first use Gaussian processes (GPs) ---well adapted to the characteristics of our input, namely temporal signals with missing data, and whose outputs are distributions over time--- to learn meaningful signals about individual hormonal dynamics. A dilated convolutional network is then trained to map from those uncertain, potentially noisy GP outputs, to target hormone levels through time. \textbf{(3)} To capture the interactions among the different hormones in a single individual through time (e.g., $hormone_i$ at $day_t$ and $hormone_j$ at $day_{t^\prime}$), we use multi-task or multi-output GPs, rather than univariate and independent GPs as experimented with in previous work. \textbf{(4)} We propose to use a a non-causal dilated convolutional neural network architecture to reconstruct personalized time-series, which yields reduced reconstruction error when compared to recurrent networks traditionally used for time series analysis. \textbf{(5)} Finally, we propose a Bayesian optimization-like approach to identify times at which to measure hormone levels that are most likely to yield accurate predictions of the proposed framework. In our set-up, where the hormonal dynamics operate at different time resolutions, yet the sampling occurs at specific times, the timing of the input samples plays a critical role on the pattern reconstruction. We describe, given a sampling budget, a way to identify the sampling times for best reconstruction of hormone levels at the population level, further understanding general hormone physiology.

A Python implementation of the proposed framework for personalized modeling of female reproductive hormonal patterns, along with an example dataset of 60 individuals, is publicly available in \href{https://github.com/iurteaga/hmc}{https://github.com/iurteaga/hmc}.

%% file: dataset.tex
Due to the lack of a publicly available gold-standard dataset of daily female reproductive hormonal measurements throughout a cycle, a synthetic dataset is for now the only alternative to design modeling techniques. Here we describe our approach to generate a dataset, which is both \textbf{diverse} ---it provides heterogeneous hormone level patterns, consistent with different types of individuals in a population--- and \textbf{realistic} ---it is grounded in real-world, high-level characteristics of the menstrual cycle. 


We leverage the previously clinically validated mechanistic model of \cite{j-Clark2003}. The model accepts a large number of parameters, each with a range of admissible values that produce heterogeneous yet realistic hormone level patterns, resulting in different cycle characteristics (\eg cycle and phase lengths). To ensure that the simulated signals reflect a range of realistic characteristics of menstrual cycles, we turned to a real-world dataset of women tracking their menstrual cycle~\citep{clue}. While menstrual trackers do not collect specific hormone levels, users can opt to track the results of their ovulation tests in addition to their cycle length. These two variables (cycle length and ovulation day) help in turn anchor different phases of the menstrual cycle. 

\begin{wrapfigure}{r}{0.45\textwidth}
	\vspace{-3ex}
	\centering
	\includegraphics[width=2.8in]{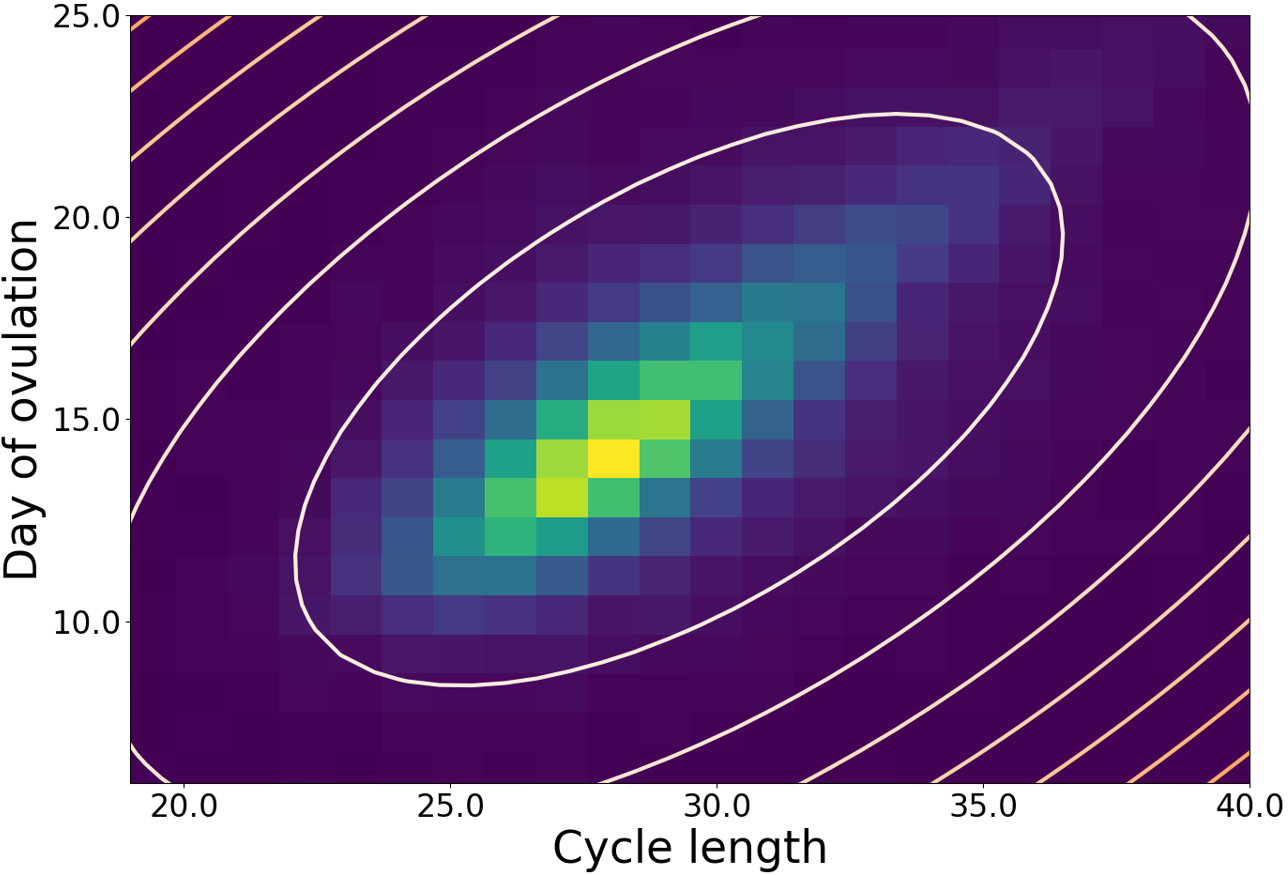}
	\vspace{-4ex}
	\caption{Density distribution of cycle length and ovulation day from a cohort of individuals aged 22-30 with natural cycles.}\label{fig:clue}
	\vspace{-5ex}
\end{wrapfigure}

We select a cohort of healthy women with natural cycles (\ie not taking any contraceptive hormonal or non-hormonal therapy) in the ages of 22-30 (when menstrual cycles are more stable during the reproductive years~\citep{treloar1967variation}) and who tracked their ovulation test results. Figure~\ref{fig:clue} shows the empirical density of these two variables of interest across the resulting cohort, for a total of 50,000 cycles. While it confirms common clinically-validated knowledge about the menstrual cycle (most likely cycle length is 29 days, and most likely ovulation day is 15), it also reflects admissible variations in the population~\citep{treloar1967variation}. Figure~\ref{fig:clue} also shows a Gaussian approximation to the observed ovulation-day/cycle-length empirical distribution, with mean $\mu= \begin{bmatrix} 15.5 & 29.1 \end{bmatrix}^T$ and covariance matrix $\Sigma=\begin{bmatrix} \begin{bmatrix}25.5 & 8.0\end{bmatrix}^T,\begin{bmatrix}8.0 & 12.6\end{bmatrix}^T\end{bmatrix}$.

To simulate a single individual's hormonal dynamics, one can first sample an ovulation-day/cycle-length pair from the Gaussian approximation of the observed empirical distribution ---ensuring that the underlying high-level characteristics of the simulated cycles are consistent with a real-world population--- and then select a random synthetic set of simulated cycles that match these characteristics (\ie an individual with realistic cycle features).

This proposed method can be used to simulate as large of a dataset as needed. Since our goal here is to identify a low-cost method to hormone level reconstruction, we aim for few individuals for which we would practically obtain gold-standard hormonal daily measurements. Therefore, in this work we simulate a small dataset of 60 diverse individuals.

%% file: methods.tex

The overall aim of our work is an end-to-end statistical framework for personalized, accurate, minimally invasive modeling of female reproductive hormonal dynamics. Specifically, our goals are to \textbf{(1)} infer and forecast individual hormone levels over time with high-fidelity; and \textbf{(2)} accommodate pragmatic and minimally invasive measurement settings. To attain accurate hormone level reconstruction with as little data as possible ---both in the number of individuals needed for training, as well as in the number of hormonal measurements required--- we devise a flexible statistical framework with two main components: a model for reconstruction of hormone levels from few measurements, and a sampling mechanism for optimal reconstruction accuracy with limited sampling budget (\ie minimally invasive requirements).

\subsection{Accurate reconstruction of hormone level dynamics}
\label{ssec:reconstruction}

With our framework, we seek to reconstruct the daily evolution of hormone levels that regulate the menstrual cycle (\ie $E$, $P$, $Ih$, $FSH$, and $LH$), from a few of their irregularly-observed measurements. We cast the problem of reconstructing these hormones over time as a \textbf{multivariate time-series regression task}. For our model to have clinical relevance, we must accommodate realistic hormone measurement practices: in essence, the less-invasive our assumptions are, the better. To that end, we aim at a very reduced and flexible measurement schedule, \ie women are asked to provide hormone levels as scarcely as possible, and not necessarily subject to a regular schedule (\eg once per week). Furthermore, all five hormone levels are attained simultaneously from the same serum-based measurement. These non-invasive, yet stringent, requirements raise several challenges for a reconstruction algorithm: available measurements will be sparse, uncertain and not uniformly sampled in time.

\begin{figure}[t]
	\vspace{-1ex}
	\centering 
	\includegraphics[width=\textwidth]{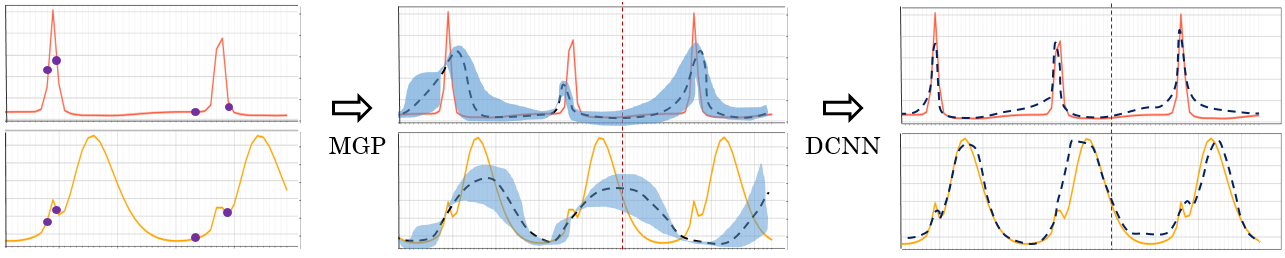} 
	\vspace{-4ex}
	\caption{Reconstruction of hormone levels. A few measurements (purple dots) are sampled from the true hormonal levels (red and yellow solid lines) of the two first cycles and constitute the input of the MGP. From this subset of points, the MGP generates a probabilistic distribution (blue area) on an extended range of time (2 original cycles and one forecasted). Samples from the MGP posterior distributions for all individuals in the cohort are then fed into a population-level DCNN, whose goal is to reconstruct the hormonal levels (blue doted line) over the three cycles.}
	\label{fig:architecture} 
	\vspace{-4ex}
\end{figure}

Figure~\ref{fig:architecture} describes our end-to-end approach to reconstruct and predict daily hormone levels. For a single individual, a Multi-task Gaussian Process (MGP) learns, given a few non-uniformly sampled measurements, a personalized posterior distribution of hormone levels over time. These hormone level distributions are subsequently used as input, over equally spaced time instants, to a Dilated Convolutional Neural Network (DCNN) whose mission is to correct the potential mispredictions of each individual MGP by learning at the population level. The DCNN incorporates general knowledge of hormonal dynamics, as it is trained over the set of MGP distributions for a cohort of individuals --- which, to ensure our pragmatic goal of low-cost and limited measurements, is of reduced size.

\subsubsection{Multi-task Gaussian Processes (MGP) for individual sparse data}
\label{sssec:mgp}

Gaussian processes~\citep{b-Rasmussen2006} are useful for time-series modeling, because they can naturally accommodate irregular and uncertain observations, and provide probabilistic predictions. We leverage their flexible and probabilistic nature to model irregularly sampled individual sparse hormone levels. Here we extend on the previous success of GPs to model reproductive hormones~\citep{j-Urteaga2017} by considering multi-task Gaussian processes, which allow for an interpretable, yet more flexible, learning of the patterns across hormone levels: both in time and across hormones.

A GP is a stochastic process such that any finite collection of random variables has a multivariate Gaussian distribution. In particular, a collection of random variables ${f(x) : x \in X }$ is said to be drawn from a GP with mean function $m(\cdot)$ and covariance function $k(\cdot, \cdot)$ if for any finite set of elements $x_1, \cdots , x_k \in X$, the associated finite set of random variables $f(x_1), \cdots, f(x_k)$, follows $f(x) \sim GP(m(\cdot), k(\cdot,\cdot))$ where $m(x) = \mathbb{E}[f(x)]$  and $k(x, x')=\mathbb{E}[(f(x)-m(x))(f(x')-m(x'))]$. A GP can be seen as a probability distribution over arbitrary functions, with $m(\cdot)$ its mean function, and $k(\cdot, \cdot)$ the covariance kernel or function.

GPs have been successfully applied to the analysis of medical time-series data, \eg~\citep{j-Stegle2008,ic-Schulam2015}. When analyzing multiple covariates over time, the MGP regression framework allows for joint modeling of multiple (irregularly sampled) time-series, as shown in several healthcare and other biomedical applications \citep{ic-Bonilla2008,ic-Alvarez2009,ip-Duerichen2014,j-Duerichen2015,j-Cheng2017}. We here consider an MGP whose input is time, and the output a five dimensional vector (one dimension or task per hormone), using a multi-output kernel. Let's consider the full set of $H$ hormone levels over $T$ time steps in matrix form $\mathbf{Y}\in \Real^{H\times T}$, where $y_{h,t}$ is the hormone $h$ level at time $t$, and $\mathbf{y}=vec{\mathbf{Y}}$ its vector counterpart. By extension of the Gaussian process principles to multivariate series, we can write the distribution of the observed hormone levels as $\textbf{y} \sim \N(\textbf{y} | \mu, \Sigma)$, with $\mu \in \Real^{H T}$ and $\Sigma \in \Real^{HT \times HT}$.


We propose to learn an MGP for each individual $i$ from the inferred hormone patterns $y_i(t_i)$, sampled at (few) time-instants, \ie $t_i \subset \{1,\cdots,T\}$. Given a set of observations, we can learn the parameters $\theta_{h}$ of the per-hormone time kernel $k^{h}(t,t^\prime)$ and the multi-output matrix $K(h,h^\prime)$ for hormones $h, h^\prime \in {1,\cdots,H}$, that maximize the marginal likelihood of the data. The output of the learned MGP is a posterior distribution over the hormone levels $z_i(t)$ at the evenly-spaced time index $t=\{1,\cdots,T\}$, \ie $z_i(t)\sim MGP(m_i(\cdot), k_i(\cdot, \cdot)|\theta_i)$. We focus on a model that shares covariance function forms on the time input, and uses a free-form positive semi-definite kernel as the covariance function across hormones. This modeling choice is flexible and has been successful for capturing (interpretable) inter-task dependencies, while avoiding the need for large training data~\citep{ic-Bonilla2008}. Mathematically, the full covariance matrix of size $HT \times HT$ follows $\Sigma = K(h,h^\prime) \otimes k(t,t^\prime)$, where $\otimes$ denotes the Kronecker product, $K(h,h^\prime)$ a positive semi-definite matrix that specifies the inter-hormone dependencies, and $k(t,t^\prime)$ the covariance matrix over time-inputs. An important property of this kernel choice is that, due to the not block-diagonal structure with regards to hormones, observations of one hormone can affect the MGP predictions of another hormone. If per-hormone \iid noise is included in the model, then we simply extend the covariance structure to accommodate a $H \times H$ diagonal matrix $\textbf{D}$ of per-task noise variance, resulting in an updated covariance structure $\textbf{y} \sim \N(\textbf{y} | \mu, \Sigma)$ where $\Sigma = K(h,h^\prime) \otimes k(t,t^\prime) + D \otimes I$.

Leveraging the recommendations of~\citet{j-Urteaga2017}, we employ the exponential periodic kernel $k(t,t^\prime)=\exp\left(\left[2 \sin^2\left(\frac{\pi}{p}|t-t^\prime|\right)\right]/l^2\right)$, with period length $p$ and lengthscale parameter $l$ to be learned; as well as a low-rank approximation to the tasks matrix through the across-hormone kernel $K(h,h^\prime)=VV^\top+diag(v)$. After learning all kernel parameters (we assume a zero-mean prior function), the MGP provides a posterior distribution over unobserved levels at any given time-instant for the hormone(s) of interest. We finally emphasize that, because of our modeling choices, the kernel function of the learned MGP provides an individualized and interpretable signature: for a given individual $i$, $k_i(t,t^\prime)$ reflects the time-varying pattern, while $K_i(h,h^\prime)$ captures the correlations between hormone levels.

For the remainder of this work, we focus on the $H=5$ hormones of interest: $E$, $P$, $Ih$, $FSH$, and $LH$. Therefore, both the observed and the inferred variables are five dimensional hormone level vectors indexed by time, \ie $y_i(t),z_i(t) \in \Real^5$, with $\theta_i$ denoting the mean and kernel parameters learned for individual $i$. We will write $y_i$, $z_i$, when it is not necessary to specify the argument (\ie the time dependence).

\subsubsection{Dilated Convolutional Neural Network (DCNN)}
\label{sssec:dcnn}

Conventional wisdom within the deep-learning community typically considers recurrent networks for time-series analysis~\citep{b-Goodfellow2016}. However, several recent results indicate that convolutional architectures can outperform recurrent networks on several sequence to sequence tasks such as audio synthesis, word-level language modeling, and machine translation~\citep{j-Oord2016, j-Kalchbrenner2016, j-Dauphin2016, j-Gehring2016, j-Gehring2017}. Aligned with these in spirit, and based on a recent comparison benchmark by~\citet{j-Bai2018}, we hereby consider convolutional networks for hormonal time-series reconstruction.

The key benefits on the use of convolutional structures for sequence modeling are \textbf{(1)} reduced gradient instability (\ie exploding and vanishing gradients) than in recurrent architectures; \textbf{(2)} ability to capture long dependencies in the input sequence via dilated layers; and \textbf{(3)} a lower memory requirement in training (especially for long input sequences), since the convolution filters are shared across layers, and the backpropagation path depends only on the network depth.

Intuitively, the idea of applying CNNs to time-series analysis is to learn convolutional filters that capture the patterns in the data. A CNN consists of a sequence of convolutional layers, the output of which is connected only to local regions of the input. This is achieved by sliding a filter $f$ --- a weight matrix --- over the input, and at each point computing the dot product between the two. A classic linear-causal convolution is only able to look back at a history proportional to the depth of the network, which limits its capability to accommodate long dependencies. To overcome this limitation, we leverage dilated convolutions and extend them to the non-causal setting: \ie the filters compute the convolution over past, present and future data. Our proposed CNN architecture is based on recent convolutional networks for sequential data~\citep{j-Oord2016,j-Lea2016}, and extends the dilated layers to the non-causal setting~\citep{j-Yu2015,j-Bai2018}: both past and future samples are used for the reconstruction of each time-instant. By employing non-causal dilated convolutions, we accommodate exponentially large receptive fields in our hormone level reconstruction. Note that the word \textit{causal} here indicates, as in signal-processing, that the filter output depends only on past and present inputs. Our proposed filters, whose output also depends on future inputs, are \textit{non-causal}.

\begin{figure}[!t]
	\centering 
	\includegraphics[width=0.5\textwidth]{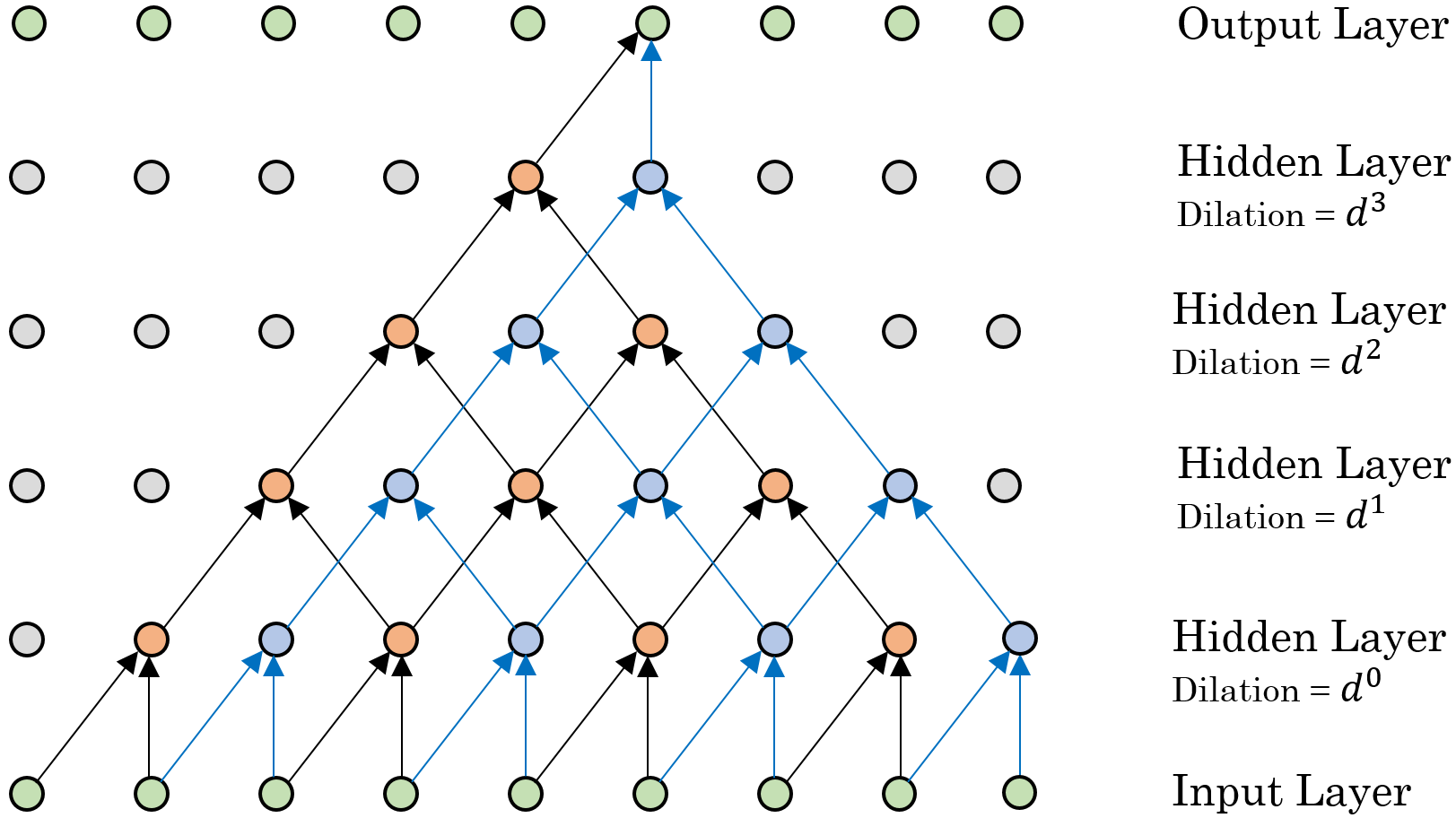} 
	\vspace{-1ex}
	\caption{Dilated Convolutional Neural Network with 4 hidden layers, $d = 2$ and $K = 2$.}
	\label{fig:Dilated_Convolution} 
	\vspace{-2ex}
\end{figure}

Mathematically, for a sequence input $X$ of size $t=\{1,\cdots,T\}$ and a filter $f$ of size $K$, the causal $d$-dilated convolution operation $F$ on element $X[t]$ of the sequence is defined as $F[t]=\left(X *_d f \right)[t]=\sum_{k=0}^{K-1} f[k] X[t-dk]$. Dilation is equivalent to introducing a fixed step between adjacent filter taps ---when $d = 1$, a dilated convolution reduces to a regular convolution. For a fixed dilation factor $d$, the outputs at the top level depend on a exponential range of inputs: specifically, the effective history for the hidden layer $l$ is $d^l\times (K - 1)$. With this architecture, for a fixed number of hidden layers, there are two degrees of freedom to increase the receptive field of the network: the filter size $K$, and the dilation factor $d$.
In this work, we extend the dilation architecture to be non-causal: \ie for each output $F[t]$, the filter is applied both backward and forward in time, centered at $t$ (see Figure~\ref{fig:Dilated_Convolution} for an illustration).

In each hidden layer $h_i$ of the proposed dilated, non-causal convolutional architecture, we use a ReLU non-linearity as our activation function, along with a residual connection, to further transform the output from the dilated convolution: \ie $h_{i+1}=ReLU\left(h_{i}+F(h_{i})\right)$. The residual connections are implemented by skipping one or more layer(s), thus adding unmodified inputs from the skipped layers to the output activation layer.

\subsubsection{Training the DCNN across a population of MGP posterior distributions}
\label{sssec:end_to_end_reconstruction}

The MGP is a full probabilistic model of the hormone levels of interest over time, which we leverage to train our proposed DCNN. We harness full potential of the generative model by drawing $S$ hormone level sample streams $z_i^{(s)}, s= \{1, \cdots, S\}$ from the learned MGP, instead of resorting to its expected value: \ie we leverage the full posterior distribution, and not only its mean prediction. The advantage of this approach is two-pronged: the DCNN is trained with uncertain inputs (\ie $S$ drawn signals $z_i^{(s)}$ are mapped to one true output $y_i$), and further, the neural network is trained to learn these mappings at the population level. By mapping all these samples to the same target function--- the true hormone  measurements--- we build a noise resistant neural network at the population level. Note that with such approach, training of a convolutional neural network is possible, even when data from few individuals is available.

Learning the DCNN given all the time-series $z_i^{(s)}, \forall (i,s)$, involves learning the parameters $w$ of the DCNN that minimize the mean-squared error at the population level. 
We optimize the DCNN based on gradient descent, in which we incrementally update the weights based on the gradient of the error function: $w(\tau + 1) = w(\tau )-\eta \nabla \mathcal{L}_2(w(\tau))$ for $\tau \in \{1, \cdots, \tau_{max}\} $ where $\tau_{max}$ is the maximum number of training iterations, and $\eta$ is the learning rate.
We compute the prediction error via the expected $\mathcal{L}_2$ norm: the distance between the DCNNs prediction $g(z_i)$ and the true hormone levels $y_i$, averaged over all individuals, \ie $\forall i\in{1,\cdots,I}$. Further, since $z_i$ is a random variable itself, we optimize the expectation of the loss function via its empirical average:
\begin{equation}
\begin{split}
w^\ast = \operatorname*{argmin}_w \sum_{i=1}^{I} \mathbb{E}_{z_i \sim MGP(\cdot|\theta_i)} \left\{\mathcal{L}_2\left[g\left(z_i, w\right), y_i\right] \right\} = \operatorname*{argmin}_w \sum_{i=1}^{I} \sum_{s=1}^{S}\left[g\left(z_i^{(s)}, w\right) - y_i\right]^2
\end{split}
\end{equation}

\subsection{Sampling for optimal hormone reconstruction via Expected Distance}
\label{ssec:sampling}

As in any other signal processing problem, the reconstruction and prediction of hormone levels hinges upon the sampling strategy of the measurements. In the context of independent GP based hormone prediction task, \citet{j-Urteaga2017} showed that sampling times influence critically the ability to predict phases accurately, and suggest that knowledge of the timing of ``peaks and valleys'' of hormone levels is needed (\eg day at which LH surge takes place for a given individual).

However, an open question remains: how to determine, given a realistic budget of measurements (\ie a reasonable number of days for a women to collect serum levels), the sampling strategy that allows for most accurate reconstruction. In this section, we describe a method to devise, given a fixed amount of possible days, which specific days are best to collect the hormone measurements. Note that, consistent with our goal of minimally invasive measurements, we assume that all five hormones are sampled simultaneously. As such, we do not look to optimize the best sampling scheme for each hormone, but rather consider optimal sampling for all hormones at once. 

The notion of selecting a subset of points for regression was first suggested by~\citet{j-Silverman1985}, and later applied via a greedy selection of points within the GP context by~\citet{ic-Smola2001}. Methods for determining accurate inducing points for sparse GP regression have been popular, \eg pseudo-inputs in ~\citep{ic-Snelson2006} and a more general view of probabilistic GP approximations in~\citep{ip-Bauer2016}.

However these strategies have been developed to approximate a full GP with as few datapoints as possible, so that the computational cost of the GP is reduced. Our goal differs as we are not working with big datasets nor do we want to reduce computational burden ---in our setting, it is feasible to compute the full posterior of the observed hormone levels.

An alternative of interest to the task at hand is that of Bayesian optimization~\citep{j-Frazier2018}, where one optimizes an objective function based on a surrogate which quantifies uncertainty via an stochastic process. By defining an appropriate acquisition function, one can then use the surrogate to decide where to sample towards the goal of maximizing the objective. Inspired by this framework in general, and by the \textit{Expected Improvement} acquisition function in particular~\citep{j-Frazier2018}, we hereby present our sampling approach for hormone reconstruction.

We propose to greedily select input points where the expected absolute difference between the true levels and the predictions of the MGP trained on the limited subset is maximized. Unlike with the expected improvement acquisition function, and more broadly any Bayesian optimization function, we are not trying to find the maximum (or minimum) of an unknown function. Instead, we greedily find the set of points that allow for a MGP with limited inputs to learn a good approximation to the true hormonal levels. That is, we find the time instant with the maximum \textbf{Expected Distance}: a function that measures the expected (over the underlying uncertainties) distance from the true hormone levels to the MGP predictions, trained over a subset of the points.

\begin{figure}[!ht]
	\vspace{-1ex}
	\begin{center}
		\includegraphics[width=0.9\textwidth]{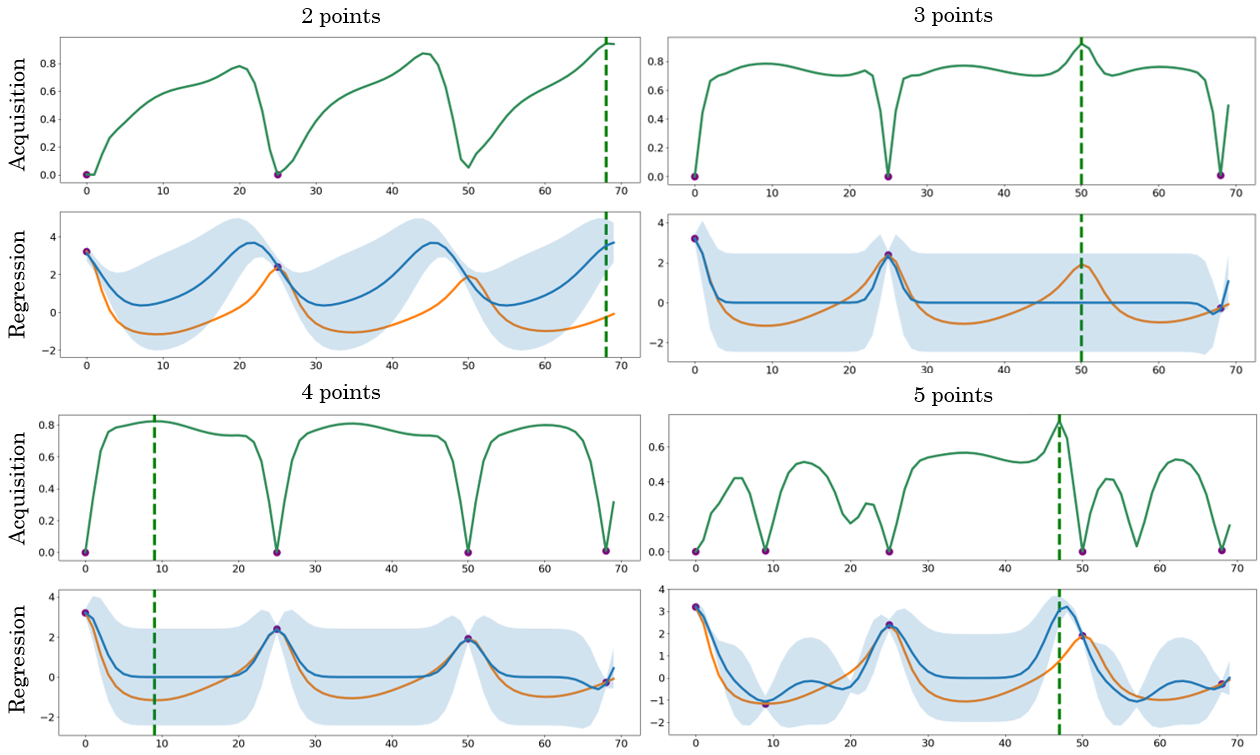}%
	\end{center}
	\vspace{-2ex}
	\caption{ED sampling with a budget of 5 days for a single hormone, starting from 2 samples: each additional sample (in purple) is identified in a greedy fashion based on the difference between, given the available sampled days, the learned posterior $z_i(\D_i)$ (in blue), and the true hormone dynamics (in orange), indicated by the green vertical line.}
	\vspace{-2ex}
	\label{fig:ED_sampling} 
\end{figure}

We define the distance function $\Psi_{i}^{h}(t|\D)= \eValue{\left|y_{i}^{h}(t)-z_{i}^{h}(t|\D)\right|}$ for hormone $h$ and individual $i$, where the expectation is computed over the randomness of the MGP output $z_i(t|\D)$. We explicitly indicate that the MGP has been trained over a subset of the data $\D=\{y(t), t\subset\{1,\cdots,T\}\}$. We find the time-instant that maximizes this expected distance with respect to the population distribution, which we approximate with the observed sample average
\begin{equation}
t^*=\argmax_{t} \eValue{\Psi_{i}^{h}(t|\D)} =\argmax_{t} \sum_{i=1}^{I} \sum_{h=1}^{H} \Psi_{i}^{h}(t|\D) \; ,
\end{equation}
where $\Psi_{i}^{h}(t|\D)$ can be computed in closed form (see details below). Because at each time instant all hormones are measured simultaneously, and we consider their accuracy equally important, a third empirical average over hormones is computed. An illustrative example of the proposed optimal sampling on a single hormone is presented in Figure~\ref{fig:ED_sampling}.

We now derive the expectation over the MGP posterior, \ie $\Psi_i(t)  = \eValue{\left|y_i(t)-z_i(t|\D)\right|}$. First, since the posterior of the MGP is parameterized by its mean $\mu_{i}(t|\D)$ and covariance $k_i(t,t^\prime|\D)$ functions, we have for any given time instant $t$, $z_i(t|\D)=\N(\cdot|\mu_{i}(t|\D), \sigma_i^2(t|\D)) = \mu_{i}(t|\D) +\sigma_i(t|\D) \epsilon$, where $\epsilon$ is a standard Gaussian variable $\epsilon \sim \N(0,1)$. In the following derivation, and for the sake of clarity, we drop the dependency on $t$ in all variables: $z_i=z_i(t|\D), \mu_{i}=\mu_i(t|\D), \sigma_i) = \sigma_i(t|\D)$, and denote with $\phi$ and $\Phi$ the probability density function and the cumulative density function of a standard Gaussian, respectively:

\small 
\vspace*{-2ex}
\begin{equation}
\begin{split}
\Psi_i & = \eValue{\left|y_i-z_i\right|} = \eValue{\left|y_i-\mu_{i}-\sigma_i \epsilon\right|} = \int_{-\infty}^{+\infty}\left|y_i-\mu_{i}-\sigma_i \epsilon\right|\phi(\epsilon)d\epsilon\\
        & = \int_{-\infty}^{\frac{y_i-\mu_{i}}{\sigma_i}} (y_i-\mu_{i}-\sigma_i \epsilon)\phi(\epsilon)d\epsilon
+\int_{\frac{y_i-\mu_{i}}{\sigma_i}}^{+\infty} (-y_i+\mu_{i}+\sigma_i\epsilon)\phi(\epsilon)d\epsilon\\
        & = (y_i-\mu_{i})\int_{-\infty}^{\frac{y_i-\mu_{i}}{\sigma_i}} \phi(\epsilon) d\epsilon
        -\sigma_i\int_{-\infty}^{\frac{y_i-\mu_{i}}{\sigma_i}} \epsilon \phi(\epsilon) d\epsilon
		+(-y_i+\mu_{i})\int_{\frac{y_i-\mu_{i}}{\sigma_i}}^{+\infty} \phi(\epsilon)d\epsilon
		+\sigma_i\int_{\frac{y_i-\mu_{i}}{\sigma_i}}^{+\infty} \epsilon\phi(\epsilon)d\epsilon\\
        & = (y_i-\mu_{i})\Phi\left(\frac{y_i-\mu_{i}}{\sigma_i}\right)
		+\sigma_i \int_{-\infty}^{\frac{y_i-\mu_{i}}{\sigma_i}} \phi^\prime (\epsilon) d\epsilon
		+(-y_i+\mu_{i})[1-\Phi\left(\frac{y_i-\mu_{i}}{\sigma_i}\right)]
		-\sigma_i\int_{\frac{y_i-\mu_{i}}{\sigma_i}}^{+\infty} \phi^\prime (\epsilon) d\epsilon \\
        & = (y_i-\mu_{i})\left[2\Phi\left(\frac{y_i-\mu_{i}}{\sigma_i}\right)-1 \right] 
		+\sigma_i \left[\phi\left(\frac{y_i-\mu_{i}}{\sigma_i}\right)-\phi\left(-\infty\right)\right]
		-\sigma_i \left[\phi\left(+\infty\right)-\phi\left(\frac{y_i-\mu_{i}}{\sigma_i}\right)\right]\\
        & = (y_i-\mu_{i})\left[2\Phi\left(\frac{y_i-\mu_{i}}{\sigma_i}\right)-1 \right] 
+2\sigma_i \phi\left(\frac{y_i-\mu_{i}}{\sigma_i}\right)
\end{split} 
\nonumber
\end{equation}
\vspace*{-2ex}
\normalsize
\begin{equation}
\Psi_i(t|\D)=\left[y_i(t)-\mu_{i}(t|\D)\right]\left[2\Phi\left(\frac{y_i(t)-\mu_{i}(t|\D)}{\sigma_i(t|\D)}\right)-1 \right] 
+2\sigma_i(t|\D) \phi\left(\frac{y_i(t)-\mu_{i}(t|\D)}{\sigma_i(t|\D)}\right)\\
\end{equation}

The classic tradeoff between exploration and exploitation is here easily interpretable. The first term counts for the exploitation, which is predominant when we are confident about our prediction (small $\sigma_i$), and the difference $(y_i-\mu_i)$ determines our decision. On the contrary, the second term increases with the uncertainty of the MGP (high $\sigma_i$), therefore favoring exploration.

%% file: evaluation.tex
\subsection{Experimental Setup}
\label{ssec:experiment_setup}

\paragraph*{Train/test datasets:} For evaluation purposes, we create a realistic synthetic dataset, that meets reasonable hormone measurement practices as well. As proposed in Section~\ref{sec:dataset}, we simulate a reduced set of individual hormone patterns with real-world cycle characteristics. Specifically, we sample $I=60$ cycle lengths from the approximation to the empirical distribution in Figure~\ref{fig:clue}, and select a random synthetic set of $H=5$ hormone signals with matching features. We draw from the marginal Gaussian approximation to the cycle length $\N\left(\mu=29.1, \sigma^2=12.6\right)$, and each associated hormone time series comprises $T=105$ days (\ie at least 3 full cycles are included for all individuals). Time-indexes are selected at random $t_i \subset \{1,T\}$ for each individual to simulate scarce, and non-necessarily regular measurement schedules, with different sampling budgets.

We randomly split the train/test population in $I_{train}=50$ (with 10 individuals kept apart for validation purposes) and $I_{test}=10$ individuals. To train the MGP, we only make use of subsampled data points in the first 2 cycles for each individual in the training set: \ie $y_i(t_i), t_i \subset \{1,\cdots,70\}, i=\{1,\cdots, I_{train}\}$. After learning each individual MGP, the DCNN is trained with the regularly spaced signals drawn from each individual MGP and the true hormone levels for the entire training population: \ie $z_i(t), y_i(t), t=\{1,\cdots,T\}, i=\{1,\cdots, I_{train}\}$. For testing purposes, the same approach is replicated, where the MGP is trained over subsampled hormone levels of each individual in the test-set ($\D_i=\{y_i(t_i) | t_i \subset \{1,\cdots,70\}, i=\{1,\cdots, I_{test}\}\}$), and the true hormone levels are only used to compute the mean squared error (MSE) across the test-population and across the five hormones: \ie $MSE_{test}=\frac{1}{I_{test}}\sum_{i=1}^{I_{test}} \left(g(z_i(\D_i))-y_i\right)^2$. 

\paragraph*{Subsampling schemes:} In alignment with our aim of accommodating minimally invasive measurements (\eg once or twice per week), we experiment with different sampling budgets of 10, 15, 25, 35 and 70 days across two cycles. Following the finding from \cite{j-Urteaga2017}, we ensure that at least two LH peaks are sampled across the two cycles (corresponding to LH surge prior to ovulation). Pragmatically, these measurements are easily obtainable for women by using urine-based ovulation tests. To assess the value of our reconstruction framework, we experiment with two sampling strategies (that augment the number of measurements beyond these two pre-determined days): $(i)$ random sampling, which selects with uniform probability a subset of days, and $(ii)$ optimal sampling, which implements the Expected Distance (ED) based strategy described in Section \ref{ssec:sampling}.

\paragraph*{Scaling:} To facilitate the training of our MGP-DCNN model, we standardize each input-feature by removing its mean and scaling to unit variance. This scaling is applied based on the training set (\ie the test set is scaled based on data seen only on the train set, to avoid any data leakage). This standardization of hormone levels helps the training process and allows for direct comparison of results across hormones.

\paragraph*{Bayesian Optimization:} To find the optimal configuration for the proposed DCNN, we tune its hyper parameters via Bayesian Optimization, using the Python~\cite{skopt_module}. The learning rate $\eta\in [0.5e^{-3}, 2e^-2]$, the number of hidden layers $L\in\{3,\cdots,6\}$, the dilation factor $d\in\{1,\cdots,3\}$, the kernel size $K\in\{2,\cdots,9\}$ (same for all hidden layers), and the number of filters (in range 5 to 12) per layer are optimized in training.

\paragraph*{Baselines and Alternative Methods:} To evaluate our proposed framework, we consider both simple baselines and alternative versions of the proposed end-to-end model:
\begin{description}
	\item[Independent GPs] This baseline replicates the work of~\cite{j-Urteaga2017}, where each hormone is analyzed with an independent GP learned with measurements within the first two cycles, and tested on the next cycle.
	\item[LSTM] A neural network with one LSTM layer (with a five-dimensional hidden representation), followed by a fully connected layer which is applied to every time instant. Because recurrent networks do not accept missing data, we impute zero-values for the missing samples in the input time-series.
	\item[MGP] This is the first component of our end-to-end model. An individual-level MGP is trained with subsampled hormone levels of the first two cycles. We learn the parameters for the across hormone kernel $K(h,h^\prime)$ and a single time correlation kernel $k(t,t^\prime)$ in the training set, and evaluate its mean prediction on the test set.
	\item[Blockwise-MGP] We impose a blockwise structure of the kernel matrix of the MGP, allowing it to learn isolated MGPs on separable groups of hormones, each of them having its own parameters. Because LH and FSH exhibit dynamics different from the other three hormones (\eg acute peaks compared to smoother peaks for the others), we train one block for $LH$ and $FSH$, and another for $E$, $P$, and $Ih$. 
	\item[Blockwise-MGP-DCNN] This is our proposed end-to-end model, where the DCNN is trained with $S=100$ samples from the blockwise MGP's posterior distribution and the full set of training points.
\end{description}

\subsection{Results}
\label{ssec:experiment_results}

\begin{figure}[!t]
	\centering
	\subfigure[Reconstructed $LH$ dynamics]{\label{fig:LH_reconstructed}\includegraphics[width=0.45\textwidth]{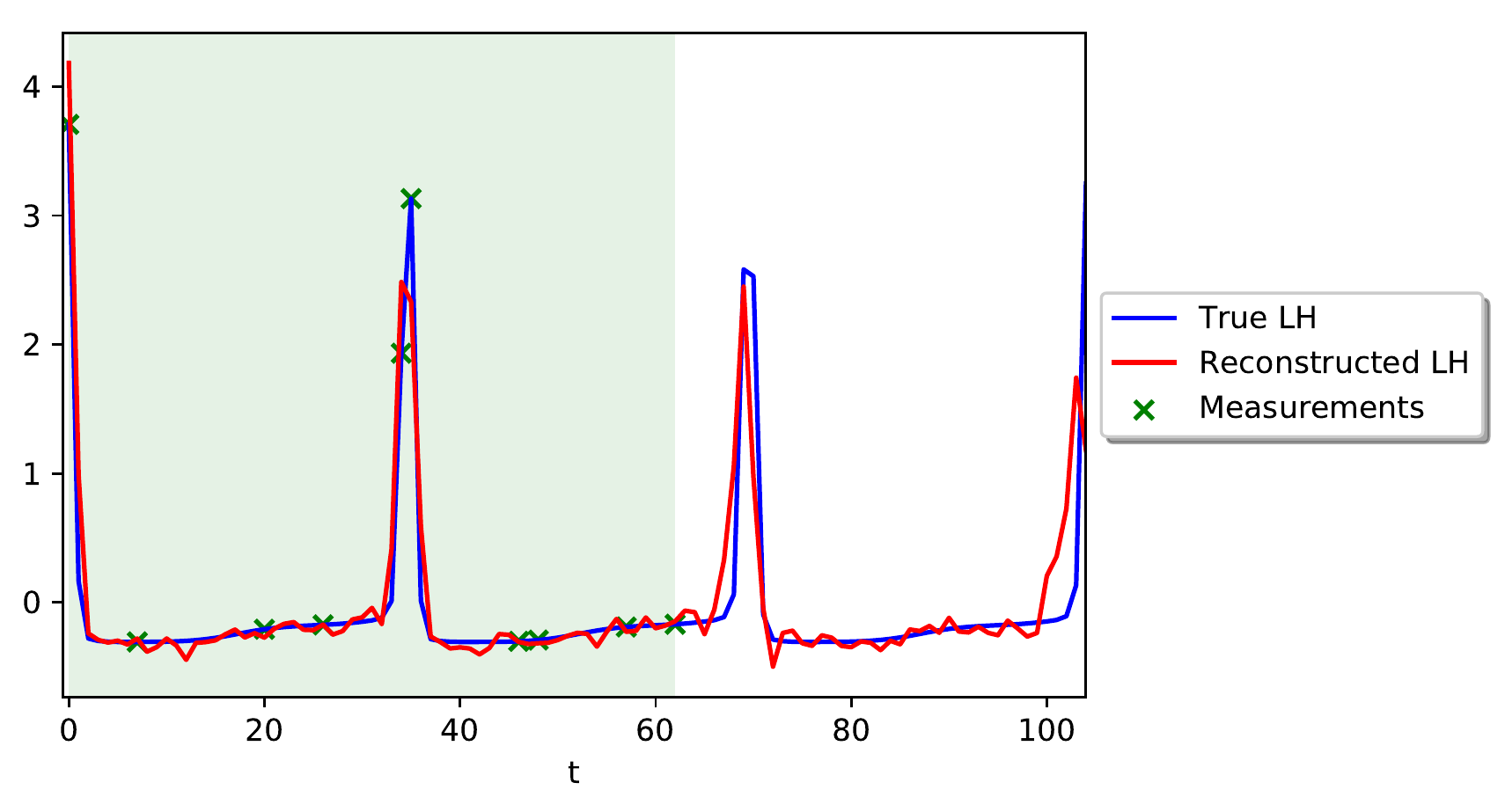}}
	\subfigure[Reconstructed $E2$ dynamics]{\label{fig:E2_reconstructed}\includegraphics[width=0.45\textwidth]{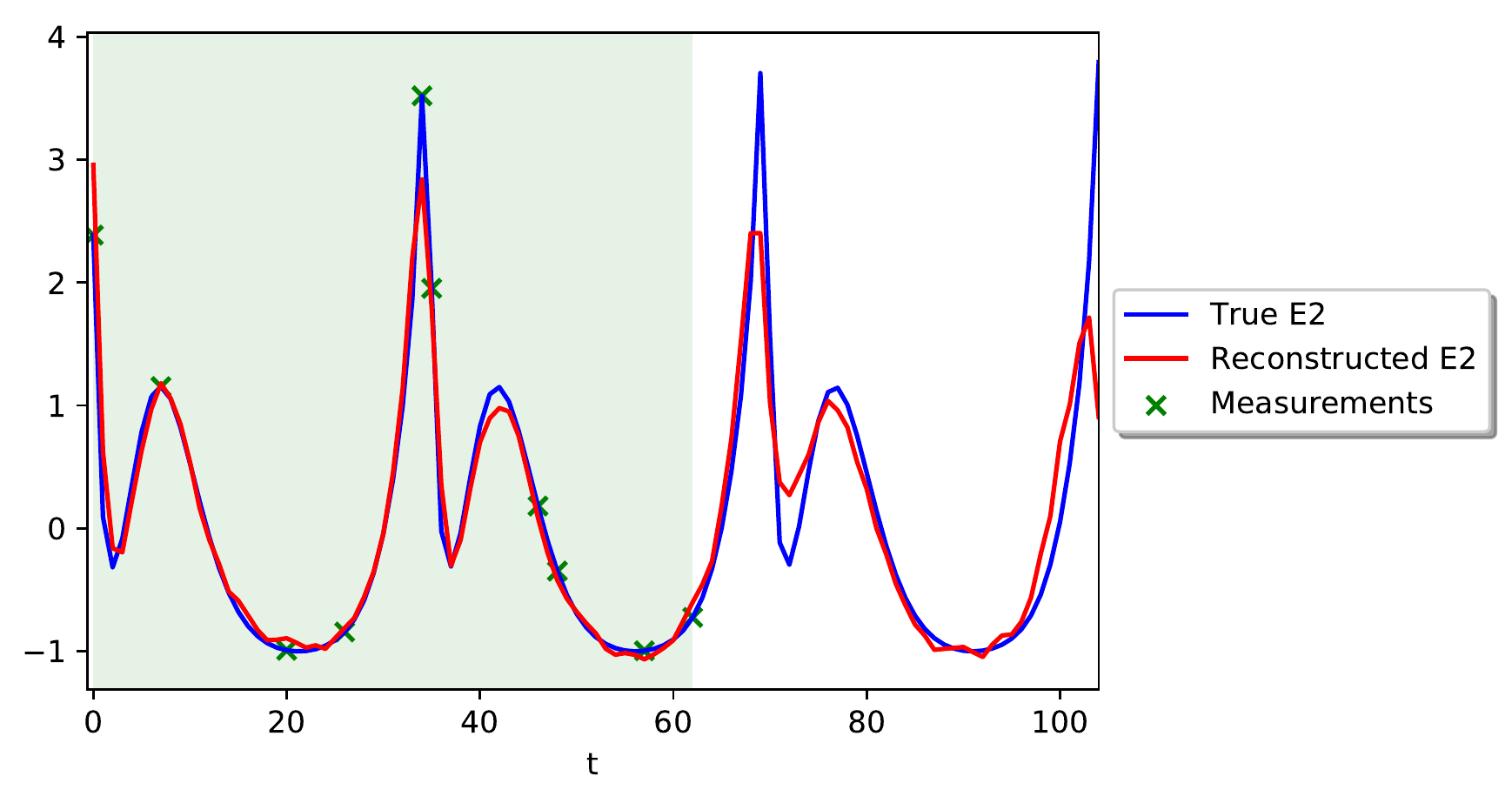}}
	\caption{Example of a MGP-DCNN based reconstruction and prediction (in red), given 10 measurements (in green), of the true hormone levels (in blue). The shaded green area indicates reconstruction time (\ie the training interval).}
	\label{fig:mgp_dcnn_reconstruction}
\end{figure}

We present in Figure~\ref{fig:mgp_dcnn_reconstruction} an illustrative example of the reconstruction and prediction of our end-to-end statistical framework: given 10 measurements within the first two cycles of an individual, the MGP-DCNN accurately reconstructs estrogen and luteinizing hormone levels across time. We further provide in Table~\ref{tab:mse_results} results for the reconstruction and prediction accuracy of all the methods described above. Note that, because we consider all the hormones to be equally important, we compute MSE results with respect to the standardized hormone levels, so that the comparison across hormones is fair \footnote{All reconstruction and prediction MSE results across all hormones, as well as per-hormone plots and MSEs are provided in Appendix~\ref{sec:evaluation_appendix}.}. 

\input{table_mse_results}

As expected, the higher the available subsampling budget is, the better the results are. In general, note that the performance with once every other day sampling budgets ($|t_i|=35$) is already satisfactory.  At lower subsampling budgets (\ie less than once every other day measurements), GP based reconstructions are less accurate and volatile, due to the uncertainty over unobserved data points, which hinder the learning of kernel hyperparameters --- specially so if only one or two measurements are available per week. As more measurements are available, the GPs outperform the LSTM, which is otherwise quite consistent across sampling budgets.

We observe that learning different temporal patterns per hormone groups is beneficial for GP based models, as well as considering correlations across them. When measurements are scarce (\ie $|t_i|=10$), learning a correlation matrix across hormones helps improve reconstruction and prediction performance. As we learn separate kernel matrices with different time-correlations per-group, temporal dependencies do not need to be the same for all the hormones, thus matching different physiological patterns.
 
The proposed framework that combines MGPs and a DCNN consistently outperforms other baselines. The reconstruction and prediction is accurate across all subsampling budgets for the Blockwise-MGP-DCNN model and, most importantly so for the lowest subsampling budget of one measurement per week. We note that, due to the randomness on the training and optimization of MGP and DCNN parameters, performance results are volatile. In our experiments, we find that the more sample streams $S$ drawn from the MGP for training the DCNN, the better and more stable the predictive accuracy of the framework is. We emphasize the two-pronged role of the MGP: ($i$) it allows training of neural nets, even with irregularly sampled time-series inputs, and ($ii$) it provides a full distribution over inputs, that allows for an uncertainty aware training of the DCNN. Further, these results support our hypothesis that the non-causal dilation allows for learning long time-horizon dependencies in our hormone level prediction task, reinforcing the claims of the literature~\citep{j-Bai2018} that convolutional architectures can be used for time-series analysis.

Finally, we note that our optimal sampling method (ED) has a strong impact on performance, both in the MGP alone and the combined MGP-DCNN setting as well. Our end-to-end framework, B-MGP-DCNN (ED), is the-best performing approach: note that optimal sampling with a budget of every other day ($|t_i|=35$, $0.037$ MSE) outperforms other methods with every day measurements ($|t_i|=70$). More importantly, B-MGP-DCNN with optimal sampling and a budget of only $|t_i|=10$ days within the first two cycles (\ie a very realistic setting of measurements every 7 days) performs better than the alternatives at high-sampling rates. 

%% file: table_mse_results.tex
\begin{table}[!t]
	\begin{center}
		\resizebox{\textwidth}{!}{
			\begin{tabular}{*{6}{|c}|}
				\hline
				\multicolumn{1}{|c|}{\multirow{1}{*}{Test-set overall MSE}} & \multicolumn{5}{c|}{Subsampling budget\cellcolor[gray]{0.7}} \\ \hline
				\cellcolor[gray]{0.7} Model & \cellcolor[gray]{0.86} $|t_i|=10$ & \cellcolor[gray]{0.86} $|t_i|=15$ & \cellcolor[gray]{0.86} $|t_i|=25$ & \cellcolor[gray]{0.86} $|t_i|=35$ & \cellcolor[gray]{0.86} $|t_i|=70$ \\ \hline
				\cellcolor[gray]{0.86}LSTM & 0.358 & 0.247 & 0.203 & 0.186 & 0.168 \\ \cline{1-6} 
				\cellcolor[gray]{0.86}Independent GPs & 1.245 & 1.066 & 0.833 & 0.140 & 0.109 \\ \cline{1-6} 
				\cellcolor[gray]{0.86}MGP & 0.823 & 1.085 & 0.708 & 0.127 & 0.057 \\ \cline{1-6} 
				\cellcolor[gray]{0.86}B-MGP & 0.818 & 1.132 & 0.682 & 0.126 & 0.119 \\ \cline{1-6} 
				\cellcolor[gray]{0.86}B-MGP (ED) & 0.723 & 0.335 & 0.513 & 0.066 & 0.115 \\ \cline{1-6} 
				\cellcolor[gray]{0.86}B-MGP-DCNN & 0.302 & 0.120 & 0.189 & 0.041 & 0.071 \\ \cline{1-6} 
				\cellcolor[gray]{0.86}B-MGP-DCNN (ED) & 0.061 & 0.045 & 0.048 & 0.037 & 0.050 \\ \cline{1-6} 
			\end{tabular}
		}
		\caption{Test-set overall average MSE for all hormones. Random sampling is used for all models, unless specifically (ED) indicated at different subsampling rates.}
		\label{tab:mse_results}
	\end{center}
	\vspace{-6ex}
\end{table}

%% file: discussion.tex
The menstrual cycle is a complex physiological process, and with access to more and more self-tracked data, computational research has very recently started to look into its modeling. Most previous work, however, has focused on high-level characteristics of the cycle, such as predicting cycle length or identifying phases based on specific signals such as body basal temperature~\citep{berglund2015identification,sosnowski2018similarity}.

To our knowledge, the work of~\cite{j-Urteaga2017} and this study are the only ones that attempt to reconstruct the full cycle, including hormone levels through time. In this work, we combine the power of generative processes (\ie MGP) with the ability of convolutional neural networks (\ie DCNN) to predict hormone levels through time, given a limited sampling budget.

Our proposed reconstruction model is inspired by and leverages several innovations in the machine learning community, yet is novel in both its goal ---accurate hormone level inference and prediction with irregular and limited budget sampling--- and its end-to-end framework ---MGP-based probabilistic training of a non-causal dilated CNN. Combining Gaussian processes, which are well adapted to datasets with uncertain and missing measurements, with neural networks has been recently explored in the context of person-level classification, both in general tasks~\citep{ip-Li2016}, as well as for sepsis prediction~\citep{futoma_learning_2017,j-Moor2019}. Our goal in this paper differs: we combine MGPs with neural nets to improved regression accuracy through time.

Furthermore, we align ourselves with a growing literature within the deep learning community~\citep{j-Oord2016, j-Kalchbrenner2016, j-Dauphin2016, j-Gehring2016, j-Gehring2017,j-Bai2018} that advocates for the use of convolutional neural nets for time-series analysis. In particular, we leverage the exponentially growing receptive fields of dilated CNNs to capture long-horizon time-dependencies, and due to the extended view that the learned MGP provides, leverage non-causal filters to improve the prediction accuracy. To the best of our knowledge, this work is novel in the specific combination of techniques (individual MGPs and a population level non-causal dilated CNN) for prediction of hormone levels in a realistic setting of limited and irregular measurements.

Finally, our work highlights the importance of time-sampling planning in the reconstruction of hormonal levels. First, inclusion of the hormone level peaks is necessary, as suggested previously by~\citet{j-Urteaga2017} and corroborated by our experiments. These peaks can be approximately derived from knowledge of a woman's cycle length, as well as via ovulation tests that directly measure $LH$ peaks, or basal-body temperature tests that do so indirectly. Second, our optimal-sampling strategy enables a minimal amount of measurements compared to random sampling. Our proposed Expected Distance function, which uses the probabilistic nature of GPs, selects the next optimal sampling time balancing the exploration-exploitation dilemma, and provides population-based insights which can be further applied in an individual-level reconstruction task. More precisely, the ED function is computed at the cohort level with respect to a normalized cycle length and, for each individual, the corresponding days of interest can be determined based on their specific cycle lengths. In practice, the optimal sampling requires pre-processing, as one would need to readjust the sampling schedule to each woman's cycle length ---this can be readily done for women with regular cycles.

In conclusion, we have proposed and validated a method to reconstruct and predict an individual's daily hormonal levels throughout the menstrual cycle based on a few hormone measurements only. This data-driven approach to reconstructing an individual's cycle patterns can help with minimally-invasive, low-cost data collection at a massive scale.

%% file: evaluation_appendix.tex
\vspace*{-2ex}
\begin{figure}[!ht]
	\centering
	\subfigure[Reconstructed $LH$ dynamics]{\label{fig:LH_reconstructed}\includegraphics[width=0.4\textwidth]{./figs/MGP_DCNN_LH_reconstruction.pdf}}

	\subfigure[Reconstructed $FSH$ dynamics]{\label{fig:FSH_reconstructed}\includegraphics[width=0.4\textwidth]{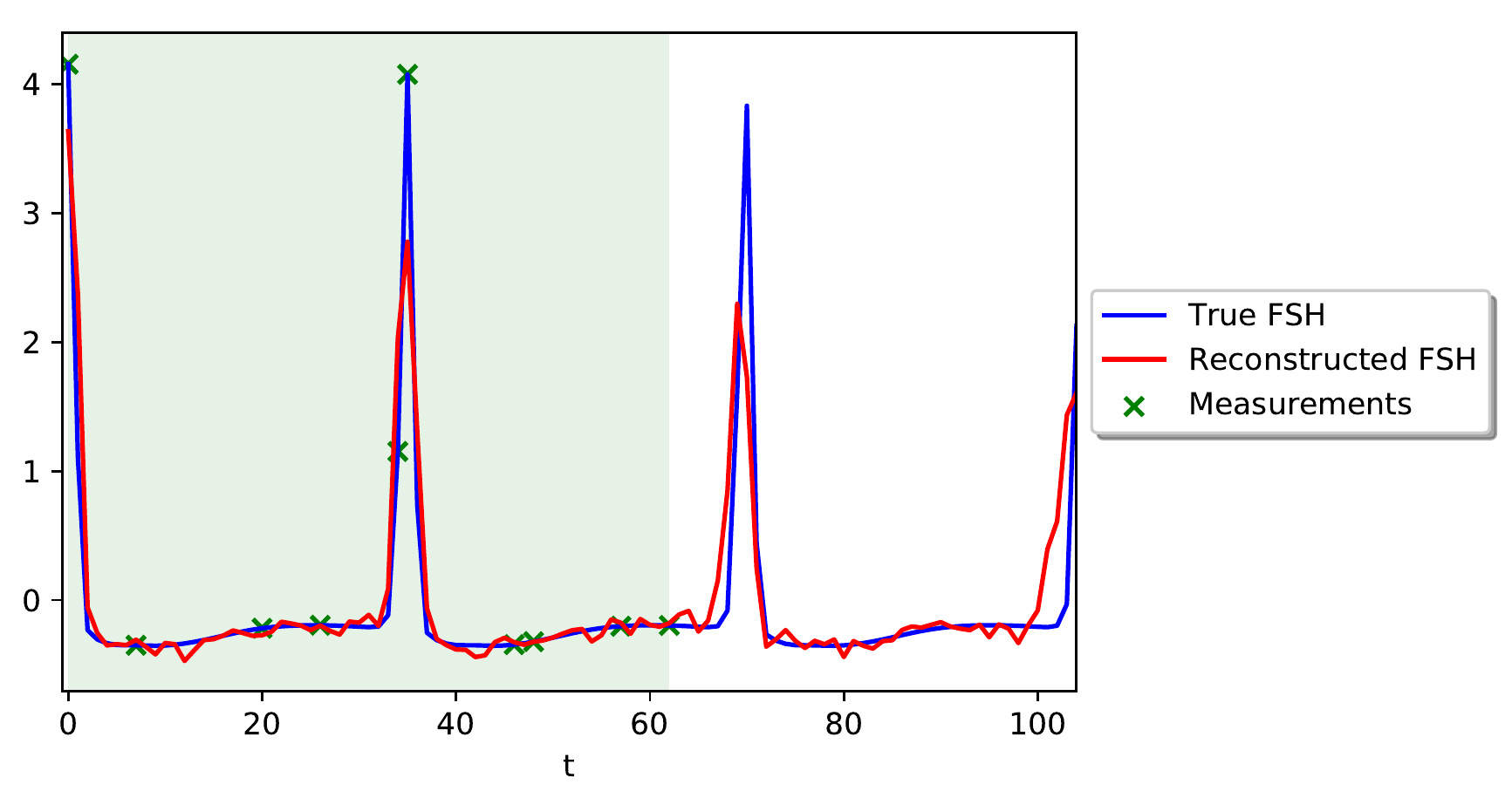}}
		
	\subfigure[Reconstructed $E$ dynamics]{\label{fig:E2_reconstructed}\includegraphics[width=0.4\textwidth]{./figs/MGP_DCNN_E2_reconstruction.pdf}}

	\subfigure[Reconstructed $P$ dynamics]{\label{fig:P4_reconstructed}\includegraphics[width=0.4\textwidth]{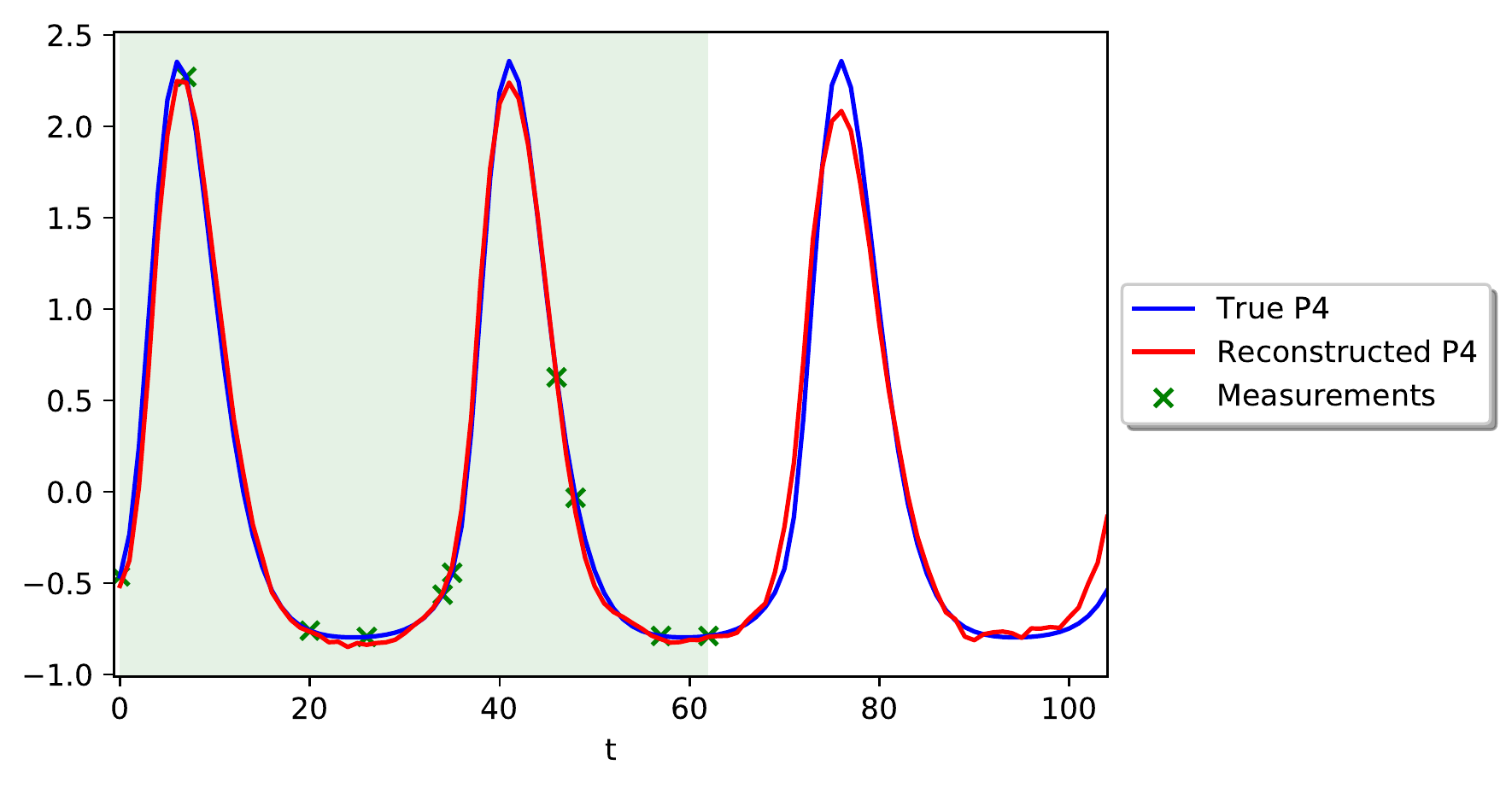}}

	\subfigure[Reconstructed $Ih$ dynamics]{\label{fig:Ih_reconstructed}\includegraphics[width=0.4\textwidth]{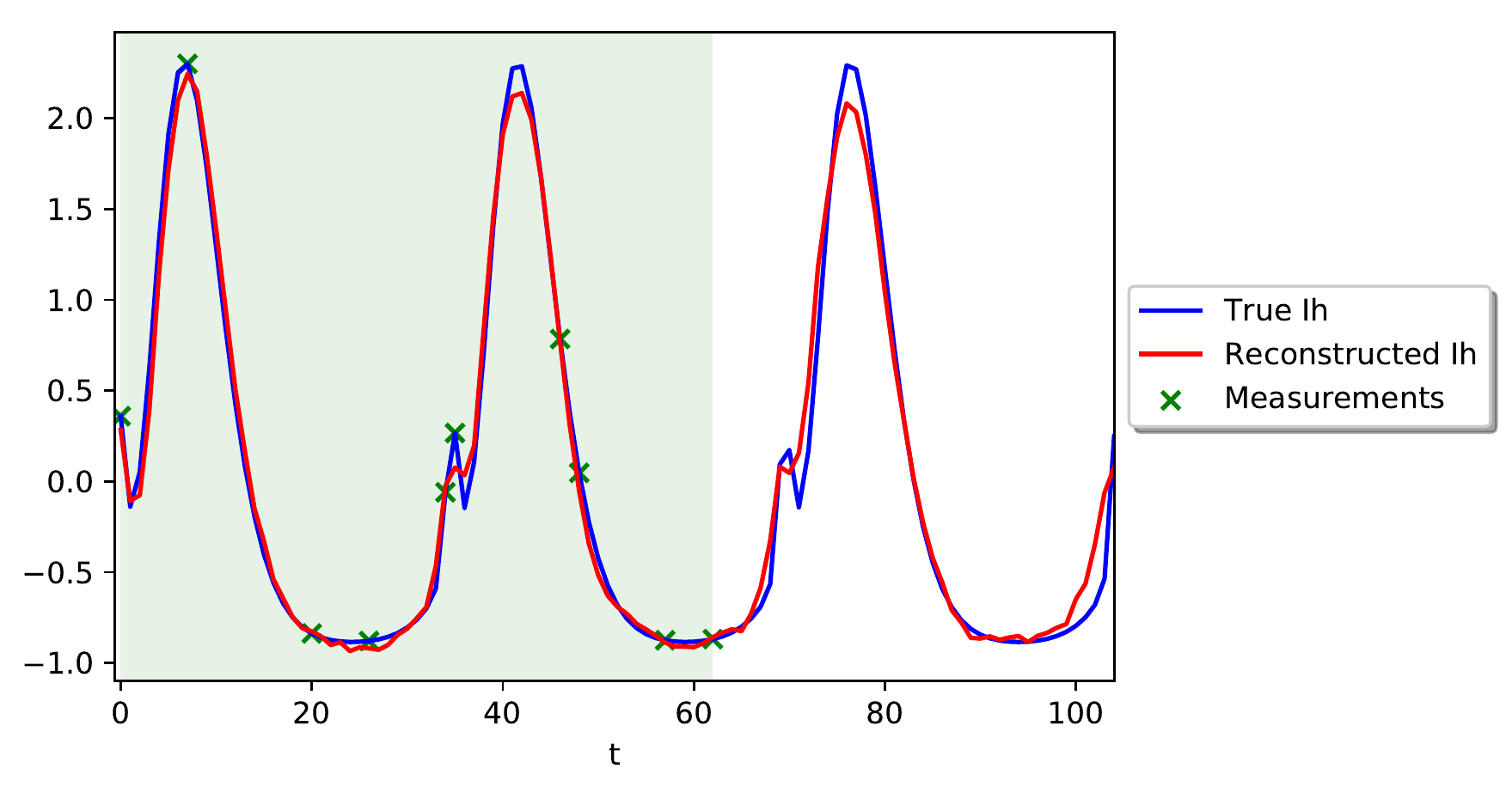}}
	\caption{Example of a MGP-DCNN based reconstruction and prediction (in red), given 10 measurements (in green), of the true hormone levels (in blue).}
	\label{fig:mgp_dcnn_reconstruction_all}
\vspace*{-2ex}
\end{figure}

\clearpage
\begin{table}[!h]
	\begin{center}
		\resizebox{\textwidth}{!}{
		\begin{tabular}{*{6}{|c}|}
			\hline
			\multicolumn{1}{|c|}{\multirow{1}{*}{Test-set reconstruction MSE}} & \multicolumn{5}{c|}{Subsampling budget\cellcolor[gray]{0.7}} \\ \hline
            \cellcolor[gray]{0.7} Model & \cellcolor[gray]{0.86} $|t_i|=10$ & \cellcolor[gray]{0.86} $|t_i|=15$ & \cellcolor[gray]{0.86} $|t_i|=25$ & \cellcolor[gray]{0.86} $|t_i|=35$ & \cellcolor[gray]{0.86} $|t_i|=70$ \\ \hline
			\cellcolor[gray]{0.86}LSTM & 0.336 & 0.240 & 0.176 & 0.173 & 0.131 \\ \cline{1-6} 
			\cellcolor[gray]{0.86}Independent GPs & 1.198 & 0.600 & 0.721 & 0.140 & 0.927 \\ \cline{1-6} 
			\cellcolor[gray]{0.86}MGP & 0.735 & 1.042 & 0.563 & 0.119 & 0.045 \\ \cline{1-6} 
			\cellcolor[gray]{0.86}B-MGP & 0.729 & 1.134 & 0.524 & 0.123 & 0.100 \\ \cline{1-6} 
			\cellcolor[gray]{0.86}B-MGP (ED) & 0.557 & 0.239 & 0.242 & 0.064 & 0.108 \\ \cline{1-6} 
			\cellcolor[gray]{0.86}B-MGP-DCNN & 0.227 & 0.096 & 0.118 & 0.035 & 0.054 \\ \cline{1-6} 
			\cellcolor[gray]{0.86}B-MGP-DCNN (ED) & 0.048 & 0.042 & 0.035 & 0.034 & 0.045 \\ \cline{1-6} 
 	    \end{tabular}
		}
		\caption{Test-set reconstruction average MSE for all hormones. Random sampling is used for all models, unless specifically (ED) indicated at different subsampling rates.}
		\label{tab:mse_results_reconstruction_all}
	\end{center}
\end{table}

\begin{table}[!h]
	\begin{center}
		\resizebox{\textwidth}{!}{
		\begin{tabular}{*{6}{|c}|}
			\hline
			\multicolumn{1}{|c|}{\multirow{1}{*}{Test-set prediction MSE}} & \multicolumn{5}{c|}{Subsampling budget\cellcolor[gray]{0.7}} \\ \hline
            \cellcolor[gray]{0.7} Model & \cellcolor[gray]{0.86} $|t_i|=10$ & \cellcolor[gray]{0.86} $|t_i|=15$ & \cellcolor[gray]{0.86} $|t_i|=25$ & \cellcolor[gray]{0.86} $|t_i|=35$ & \cellcolor[gray]{0.86} $|t_i|=70$ \\ \hline
			\cellcolor[gray]{0.86}LSTM & 0.402 & 0.261 & 0.258 & 0.213 & 0.241 \\ \cline{1-6} 
			\cellcolor[gray]{0.86}Independent GPs & 1.338 & 0.659 & 1.057 & 0.141 & 0.875 \\ \cline{1-6} 
			\cellcolor[gray]{0.86}MGP & 0.998 & 1.171 & 0.998 & 0.143 & 0.081 \\ \cline{1-6} 
			\cellcolor[gray]{0.86}B-MGP & 0.998 & 1.126 & 0.998 & 0.131 & 0.125 \\ \cline{1-6} 
			\cellcolor[gray]{0.86}B-MGP (ED) & 1.055 & 0.250 & 1.055 & 0.070 & 0.129 \\ \cline{1-6} 
			\cellcolor[gray]{0.86}B-MGP-DCNN & 0.453 & 0.134 & 0.329 & 0.049 & 0.106 \\ \cline{1-6} 
			\cellcolor[gray]{0.86}B-MGP-DCNN (ED) & 0.088 & 0.050 & 0.074 & 0.043 & 0.062 \\ \cline{1-6} 
	    \end{tabular}
	    }
		\caption{Test-set prediction average MSE for all hormones. Random sampling is used for all models, unless specifically (ED) indicated at different subsampling rates.}
		\label{tab:mse_results_prediction_all}
	\end{center}
\end{table}

\begin{table}[!h]
	\begin{center}
		\resizebox{\textwidth}{!}{
		\begin{tabular}{*{6}{|c}|}
			\hline
			\multicolumn{1}{|c|}{\multirow{1}{*}{Test-set overall MSE for $LH$}} & \multicolumn{5}{c|}{Subsampling budget\cellcolor[gray]{0.7}} \\ \hline
            \cellcolor[gray]{0.7} Model & \cellcolor[gray]{0.86} $|t_i|=10$ & \cellcolor[gray]{0.86} $|t_i|=15$ & \cellcolor[gray]{0.86} $|t_i|=25$ & \cellcolor[gray]{0.86} $|t_i|=35$ & \cellcolor[gray]{0.86} $|t_i|=70$ \\ \hline
			\cellcolor[gray]{0.86}LSTM & 0.703 & 0.475 & 0.406 & 0.389 & 0.348 \\ \cline{1-6} 
			\cellcolor[gray]{0.86}Independent GPs & 1.671 & 1.220 & 0.930 & 0.382 & 1.002 \\ \cline{1-6} 
			\cellcolor[gray]{0.86}MGP & 0.671 & 2.845 & 0.560 & 0.282 & 0.135 \\ \cline{1-6} 
			\cellcolor[gray]{0.86}B-MGP & 0.674 & 1.775 & 0.548 & 0.295 & 0.261 \\ \cline{1-6} 
			\cellcolor[gray]{0.86}B-MGP (ED) & 0.563 & 0.485 & 0.419 & 0.186 & 0.282 \\ \cline{1-6} 
			\cellcolor[gray]{0.86}B-MGP-DCNN & 0.542 & 0.196 & 0.336 & 0.071 & 0.132 \\ \cline{1-6} 
			\cellcolor[gray]{0.86}B-MGP-DCNN (ED) & 0.105 & 0.072 & 0.086 & 0.046 & 0.078 \\ \cline{1-6} 
        \end{tabular}
    	}
		\caption{Test-set overall average MSE for $LH$. Random sampling is used for all models, unless specifically (ED) indicated at different subsampling rates.}
		\label{tab:mse_results_overall_LH}
	\end{center}
\end{table}

\begin{table}[!h]
	\begin{center}
		\resizebox{\textwidth}{!}{
		\begin{tabular}{*{6}{|c}|}
			\hline
			\multicolumn{1}{|c|}{\multirow{1}{*}{Test-set overall MSE for $FSH$}} & \multicolumn{5}{c|}{Subsampling budget\cellcolor[gray]{0.7}} \\ \hline
            \cellcolor[gray]{0.7} Model & \cellcolor[gray]{0.86} $|t_i|=10$ & \cellcolor[gray]{0.86} $|t_i|=15$ & \cellcolor[gray]{0.86} $|t_i|=25$ & \cellcolor[gray]{0.86} $|t_i|=35$ & \cellcolor[gray]{0.86} $|t_i|=70$ \\ \hline
			\cellcolor[gray]{0.86}LSTM & 0.583 & 0.431 & 0.324 & 0.289 & 0.272 \\ \cline{1-6} 
			\cellcolor[gray]{0.86}Independent GPs & 1.564 & 0.900 & 0.933 & 0.228 & 0.965 \\ \cline{1-6} 
			\cellcolor[gray]{0.86}MGP & 0.772 & 1.587 & 0.699 & 0.205 & 0.085 \\ \cline{1-6} 
			\cellcolor[gray]{0.86}B-MGP & 0.769 & 2.189 & 0.671 & 0.260 & 0.188 \\ \cline{1-6} 
			\cellcolor[gray]{0.86}B-MGP (ED) & 0.569 & 0.356 & 0.426 & 0.107 & 0.209 \\ \cline{1-6} 
			\cellcolor[gray]{0.86}B-MGP-DCNN & 0.494 & 0.164 & 0.297 & 0.055 & 0.107 \\ \cline{1-6} 
			\cellcolor[gray]{0.86}B-MGP-DCNN (ED) & 0.084 & 0.051 & 0.066 & 0.037 & 0.060 \\ \cline{1-6} 
        \end{tabular}
    	}
		\caption{Test-set overall average MSE for $FSH$. Random sampling is used for all models, unless specifically (ED) indicated at different subsampling rates.}
		\label{tab:mse_results_overall_FSH}
	\end{center}
\end{table}

\begin{table}[!h]
	\begin{center}
		\resizebox{\textwidth}{!}{
		\begin{tabular}{*{6}{|c}|}
			\hline
			\multicolumn{1}{|c|}{\multirow{1}{*}{Test-set overall MSE for $E$}} & \multicolumn{5}{c|}{Subsampling budget\cellcolor[gray]{0.7}} \\ \hline
            \cellcolor[gray]{0.7} Model & \cellcolor[gray]{0.86} $|t_i|=10$ & \cellcolor[gray]{0.86} $|t_i|=15$ & \cellcolor[gray]{0.86} $|t_i|=25$ & \cellcolor[gray]{0.86} $|t_i|=35$ & \cellcolor[gray]{0.86} $|t_i|=70$ \\ \hline
			\cellcolor[gray]{0.86}LSTM & 0.319 & 0.205 & 0.160 & 0.132 & 0.105 \\ \cline{1-6} 
			\cellcolor[gray]{0.86}Independent GPs & 1.136 & 0.731 & 0.779 & 0.077 & 0.927 \\ \cline{1-6} 
			\cellcolor[gray]{0.86}MGP & 0.786 & 0.638 & 0.706 & 0.099 & 0.049 \\ \cline{1-6} 
			\cellcolor[gray]{0.86}B-MGP & 0.809 & 0.860 & 0.720 & 0.060 & 0.074 \\ \cline{1-6} 
			\cellcolor[gray]{0.86}B-MGP (ED) & 0.809 & 0.170 & 0.646 & 0.031 & 0.069 \\ \cline{1-6} 
			\cellcolor[gray]{0.86}B-MGP-DCNN & 0.295 & 0.115 & 0.188 & 0.041 & 0.073 \\ \cline{1-6} 
			\cellcolor[gray]{0.86}B-MGP-DCNN (ED) & 0.073 & 0.049 & 0.056 & 0.051 & 0.064 \\ \cline{1-6} 
        \end{tabular}
    	}
		\caption{Test-set overall average MSE for $E$. Random sampling is used for all models, unless specifically (ED) indicated at different subsampling rates.}
		\label{tab:mse_results_overall_E}
	\end{center}
\end{table}

\begin{table}[!h]
	\begin{center}
		\resizebox{\textwidth}{!}{
		\begin{tabular}{*{6}{|c}|}
			\hline
			\multicolumn{1}{|c|}{\multirow{1}{*}{Test-set overal MSE for $P$}} & \multicolumn{5}{c|}{Subsampling budget\cellcolor[gray]{0.7}} \\ \hline
            \cellcolor[gray]{0.7} Model & \cellcolor[gray]{0.86} $|t_i|=10$ & \cellcolor[gray]{0.86} $|t_i|=15$ & \cellcolor[gray]{0.86} $|t_i|=25$ & \cellcolor[gray]{0.86} $|t_i|=35$ & \cellcolor[gray]{0.86} $|t_i|=70$ \\ \hline
			\cellcolor[gray]{0.86}LSTM & 0.098 & 0.053 & 0.048 & 0.060 & 0.053 \\ \cline{1-6} 
			\cellcolor[gray]{0.86}Independent GPs & 0.918 & 0.091 & 0.757 & 0.002 & 0.800 \\ \cline{1-6} 
			\cellcolor[gray]{0.86}MGP & 0.938 & 0.146 & 0.794 & 0.020 & 0.009 \\ \cline{1-6} 
			\cellcolor[gray]{0.86}B-MGP & 0.920 & 0.395 & 0.736 & 0.003 & 0.009 \\ \cline{1-6} 
			\cellcolor[gray]{0.86}B-MGP (ED) & 0.831 & 0.085 & 0.528 & 0.001 & 0.007 \\ \cline{1-6} 
			\cellcolor[gray]{0.86}B-MGP-DCNN & 0.089 & 0.033 & 0.058 & 0.014 & 0.020 \\ \cline{1-6} 
			\cellcolor[gray]{0.86}B-MGP-DCNN (ED) & 0.020 & 0.024 & 0.014 & 0.025 & 0.023 \\ \cline{1-6} 
        \end{tabular}
    	}
		\caption{Test-set overall average MSE for $P$. Random sampling is used for all models, unless specifically (ED) indicated at different subsampling rates.}
		\label{tab:mse_results_overall_P}
	\end{center}
\end{table}

\begin{table}[!h]
	\begin{center}
		\resizebox{\textwidth}{!}{
		\begin{tabular}{*{6}{|c}|}
			\hline
			\multicolumn{1}{|c|}{\multirow{1}{*}{Test-set overall MSE for $Ih$}} & \multicolumn{5}{c|}{Subsampling budget\cellcolor[gray]{0.7}} \\ \hline
            \cellcolor[gray]{0.7} Model & \cellcolor[gray]{0.86} $|t_i|=10$ & \cellcolor[gray]{0.86} $|t_i|=15$ & \cellcolor[gray]{0.86} $|t_i|=25$ & \cellcolor[gray]{0.86} $|t_i|=35$ & \cellcolor[gray]{0.86} $|t_i|=70$ \\ \hline
			\cellcolor[gray]{0.86}LSTM & 0.086 & 0.073 & 0.079 & 0.062 & 0.061 \\ \cline{1-6} 
			\cellcolor[gray]{0.86}Independent GPs & 0.935 & 0.156 & 0.767 & 0.013 & 0.853 \\ \cline{1-6} 
			\cellcolor[gray]{0.86}MGP & 0.948 & 0.211 & 0.781 & 0.028 & 0.009 \\ \cline{1-6} 
			\cellcolor[gray]{0.86}B-MGP & 0.920 & 0.439 & 0.736 & 0.010 & 0.012 \\ \cline{1-6} 
			\cellcolor[gray]{0.86}B-MGP (ED) & 0.843 & 0.117 & 0.546 & 0.004 & 0.010 \\ \cline{1-6} 
			\cellcolor[gray]{0.86}B-MGP-DCNN & 0.091 & 0.038 & 0.063 & 0.016 & 0.023 \\ \cline{1-6} 
			\cellcolor[gray]{0.86}B-MGP-DCNN (ED) & 0.023 & 0.027 & 0.017 & 0.028 & 0.027 \\ \cline{1-6} 
        \end{tabular}
    	}
		\caption{Test-set overall average MSE for $Ih$. Random sampling is used for all models, unless specifically (ED) indicated at different subsampling rates.}
		\label{tab:mse_results_overall_Ih}
	\end{center}
\end{table}